# On the computational analysis of the genetic algorithm for attitude control of a carrier system

Hadi Jahanshahi[1], Naeimeh Najafizadeh Sari[1]

[1] Dept. of Aerospace Engineering, Univ. of Tehran, Tehran, Iran

## Abstract

This paper intends to cover three main topics. First, a fuzzy-PID controller is designed to control the thrust vector of a launch vehicle, accommodating a CanSat. Then, the genetic algorithm (GA) is employed to optimize the controller performance. Finally, through adjusting the algorithm parameters, their impact on the optimization process is examined. In this regard, the motion vector control is programmed based on the governing dynamic equations of motion for payload delivery in the desired altitude and flight-path angle. This utilizes one single input and one preferential fuzzy inference engine, where the latter acts to avoid the system instability in large angles for the thrust vector. The optimization objective functions include the deviations of the thrust vector and the system from the equilibrium state, which must be met simultaneously. Sensitivity analysis of the parameters of the genetic algorithm involves examining nine different cases and discussing their impact on the optimization results.

## 1. Introduction

Due to costly space projects, affordable flight models and test prototypes are of incomparable importance in academic and research applications, such as data acquisition and subsystems testing. In this regard, CanSat could be used as a low-cost, high-tech, and light-weight model; this makes it popular in academia [1]. CanSat is constituted from the words "can" and "sat," which collectively means a satellite that is embeddable in a soda can [2]. In these apparatuses, an electronic payload placed into a container dimensionally comparable to a soda can; it is then launched into space with a rocket or balloon [3]. The attained altitude is a few thousand meters, which is much lower than the altitude of sounding rockets [4].

The concept of fuzzy logic was introduced by Zadeh in 1965; it has been improved by several researchers, forming a potent tool for a variety of applications [5]. For example, Precup and Hellendoorn [6] and Larsen [7] have used fuzzy logic in controllers for various industrial and research applications. The control area has attracted the most significant studies on fuzzy systems [8–17]. Petrov et al. have used fuzzy-PID controllers to control systems with different nonlinear terms [18]. Hu and colleagues proposed a new and simple method for fuzzy-PID controller design based on fuzzy logic and GA-based optimization [19]. Juang et al. have used triangular membership functions in fuzzy inference systems along with a genetic algorithm to tune parameters or fuzzy-PID controllers [20]. Operating fuzzy-PID controllers and online adjustment of fuzzy parameters were the main output of Resnick et al. researches [21].

In 1950, Alan Turing proposed a "learning machine" which would parallel the principles of evolution [22]. Genetic algorithms (GAs) are stochastic global search and optimization methods that mimic the metaphor of natural biological evolution [23]. GAs consider the principle of survival of the fittest to produce better generations out of a population. Although genetic algorithms cannot always provide the optimal solution, it has its own advantages [24] and is a powerful tool for solving complex problems. GA is an effective strategy and had successfully been used in the offline control of systems by a number of studies. Krishnakumar and Goldberg [25] have shown the efficiency of genetic optimization methods in deriving controller structures in aerospace applications compared to traditional methods such as LQR and Powell's gain set design. Porter and Mohamed [26] have taken initiative and by the use of GA have offered a simple and applicable eigenstructure assignment solution which is applied to the design of multivariable flight-control system of an aircraft. Others have denoted how to use GA to choose control structures [27].

Heuristic methods are highly dependent on their agents and parameters. Therefore, GA properties (mainly population size and crossover ratio) are of high importance in finding optimum points which are usually found by sensitivity analysis. These parameters are defined for a better acquaintance of readers in the following.

This paper focuses on designing a GA-based fuzzy-PID controller. A two-termed cost function containing path and thrust vector deviations is fed into GA code to be optimized. The code adjusts the parameters. Nine different combinations with relative optimality are discussed. The paper is dissected into following sections:

- "CanSat carrier system" which presents a simple model of the carrier system

- "Fuzzy-PID controller" that describes controller design and its parameters
- "Optimization" which describes the optimization process
- "Results and discussion" that clarify results and comparisons
- "Conclusion"
- "References"

## 2. CanSat carrier system

The dynamic equations of a CanSat carrier system is derived from the Newtonian law. It should be added that in the separation stage, the projection of satellite velocity vector must be tangent to the horizontal plane. Figure 1 shows a simplified model of a launch vehicle in which $\theta$ is the angle of the longitudinal vector of the vehicle in the perpendicular direction (toward the ground) and $\varphi$ is the angle of its thrust with body centerline. The dynamics of the system can be summarized in:

$$\sum M_{CM} = I\alpha \tag{1}$$

in which $M_{CM}$ is the moment around the center of mass, $I$ is the inertial moment, and $\alpha$ is the angular acceleration about an axis perpendicular to the plane. Eq. (1) can be expanded to (2)

$$\frac{l}{2} \times F_n = I\ddot{\theta} \tag{2}$$

In the notation $l$ is used for the length of the vehicle, $F$ for the thrust force, and $F_n$ for its projection perpendicular to the longitudinal direction of launch vehicle. It is known that the vehicle moves along the vertical axis with acceleration of. Therefore, Newton equation for that axis is rearranged as below:

$$\sum F_z = ma \tag{3}$$

in which $F_z$ and $m$ are, respectively, the force along the vertical axis and the mass of the launch vehicle. Eq. (3) can be rewritten as below:

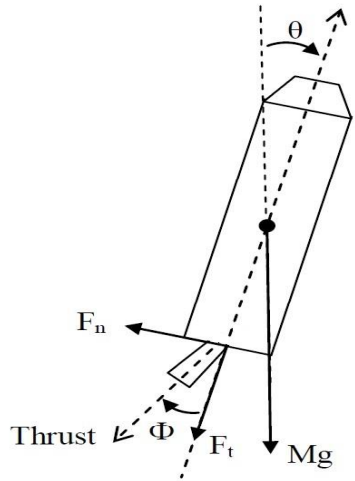

Figure 1. Carrier system scheme.

$$-mg + F_t \cos\theta = ma \tag{4}$$

Meanwhile, geometric relations dictate the following equations in the vertical plane:

$$F_n = F\sin(\varphi) \tag{5}$$
$$F_t = F\cos(\varphi) \tag{6}$$

By substituting (6) in (4), we have:

$$F = \frac{-m(a+g)}{\cos(\varphi)\cos(\theta)} \tag{7}$$

Insertion of (5) into (2) in a similar pattern yields to

$$\ddot{\theta} = \frac{1}{2I} lF\sin(\varphi) \tag{8}$$

with considering $\theta = 0$ and substitution of (7) in (8) results in:

$$\ddot{\theta} = \frac{-1}{2I}ml(a+g)tan(\varphi) \qquad (9)$$

By substituting $\frac{-1}{2I}ml(a+g)$ by $b$ and $tan(\varphi)$ by $u_t$, the dynamic equation of the system leads to

$$\ddot{\theta} = bu_t \qquad (10)$$

in which $u_t$ is the control parameter. Therefore, equations of system states take the following form:

$$\begin{aligned}\dot{x}_1(t) &= x_2(t)\\ \dot{x}_2(t) &= bu_t\\ y(t) &= x_1(t)\end{aligned} \qquad (11)$$

where $\theta$ and $\dot{\theta}$ are, respectively, represented by $x_1(t)$ and $x_2(t)$. The measurable state vector is notated by $X = [x_1, x_2]^T$.

## 3. Fuzzy controller design

Two types of fuzzy inference engines are utilized in the proposed fuzzy controller [17]. The first type is single input fuzzy inference engine (SIFIE). The second inference motor type is the preferential fuzzy inference engine (PFIE) that represents the control priority order of each norm block output.

$$SIFIM - i\ :\ \{R_i^j: if\ x_i = A_i^j\ then\ u_i = C_i^j\}_{j=1}^{m} \qquad (12)$$

The SIFIE-$i$ represents to single input inference engine which accepts the $i^{th}$ input, and $R_i^j$ is the $j^{th}$ rule of the $i^{th}$ single input inference engine. Also, $A_i^j$ and $C_i^j$ are relevant membership functions. Each input item usually has a different role in the implementation of control. In order to express the different impacts of implementing each input item in the system, single input fuzzy inference engine defines a dynamic importance degree ($w_i^D$) for each input item as (13).

$$w_i^D = w_i + B_i \times \Delta w_i \qquad (13)$$

where $w_i$, $B_i$, and $\Delta w_i$ are control parameters described by fuzzy rules. $SIFIM - i$ block calculates $f_i$ as follows:

$$f_i = \frac{NB_i \times f_1 + Z_i \times f_2 + PB_i \times f_3}{NB_i + Z_i + PB_i} \qquad (14)$$

The membership functions of SIFIEs are shown in Figure 2. As mentioned before $f_1$, $f_2$ and $f_3$, the SIFIEs fuzzy rules are extracted from Table 1.

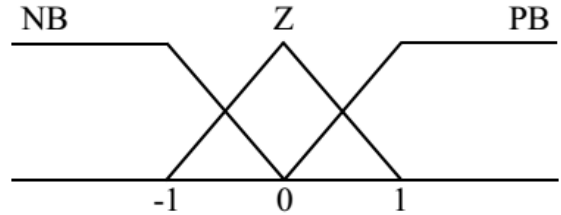

Figure 2. Membership functions of SIFIEs (note: NB = negative big; Z = zero; PB = positive big)

| If | then |
|---|---|
| $NB_i$ | $f_1 = 1$ |
| $Z_i$ | $f_2 = 0$ |
| $PB_i$ | $f_3 = -1$ |

Table 1. Fuzzy rules of SIFIEs.

The other type of fuzzy inference engine (PFIE) guarantees satellite control system performance using desired values in one or more axes of the coordinate system. PFIE-$i$ calculates $\Delta w_i$ as follows:

$$\Delta w_1 = \Delta w_2 = \Delta w_3 = \frac{w_1 \times DS + w_2 \times DM + w_3 \times DL}{DS + DM + DL} \qquad (15)$$

The membership functions of PFIEs are shown in Figure 3, while their fuzzy rules are tabulated in Table 2.

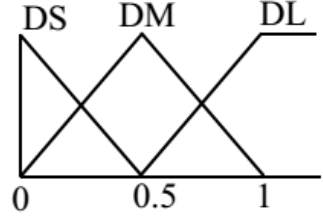

Figure 3. Membership functions of PFIE (note: DS = distance short; DM = distance medium; DL = distance long)

| If | | Then |
|---|---|---|
| | DS | $w_1 = 1$ |
| $|\theta|$ | DM | $w_2 = 0.5$ |
| | DL | $w_3 = 1$ |

Table 2. Fuzzy rules of PFIEs.

By calculating $f_i$ and $\Delta w_i$, it is possible to define fuzzy-PID controller as (16).

$$u_{fuzzy-PID} = \hat{K}_{i\theta} \int \hat{\theta} dt + \hat{K}_{p\theta} \hat{\theta} + \hat{K}_{d\theta} \frac{d\hat{\theta}}{dt} \tag{16}$$

where $u_{Fuzzy-PID}$ is the control action and $\int \hat{\theta} dt$, $\hat{\theta}$ and $\frac{d\hat{\theta}}{dt}$ are, respectively, the fuzzy forms of $\int \theta dt$, $\theta$, and $\frac{d\theta}{dt}$, and should be obtained from SIFIE. In other words, we have $\int \hat{\theta} dt = f_1$, $\hat{\theta} = f_2$ and $\frac{d\hat{\theta}}{dt} = f_3$. Parameters of $\hat{K}_{i\theta}$, $\hat{K}_{p\theta}$ and $\hat{K}_{d\theta}$ in (7) are fuzzy variables calculated by following equations:

$$\hat{K}_{i\theta} = K_{i\theta}^b + K_{i\theta}^r \Delta W_1 \tag{17}$$
$$\hat{K}_{p\theta} = K_{p\theta}^b + K_{p\theta}^r \Delta W_2 \tag{18}$$
$$\hat{K}_{d\theta} = K_{d\theta}^b + K_{d\theta}^r \Delta W_3 \tag{19}$$

in which $K_{i\theta}^b$, $K_\theta^b$ and $K_{d\theta}^b$ are the base variables and $K_{i\theta}^r$, $K_\theta^r$, and $K_{d\theta}^r$ are regulation variables. While it is possible to find these variables by try and error, the best way to find them is using optimization approaches like evolutionary algorithms, especially genetic algorithm (GA).

## 4. Optimization

GA is an approach for solving optimization problems based on biological evolution via repeatedly modifying a population of individual solutions. At each level, individuals are chosen randomly from the current population (as parents) and then employed to produce the children for the next generation. In this paper, the following operators are implemented for optimization of the fuzzy-PID controller:

- **Population size (PS):** Increasing the population size enables GA to search more points and thereby obtain a better result. However, the larger the population size, the longer it takes for the GA to compute each generation.
- **Crossover options:** Crossover options specify how GA combines two individuals, or parents, to form a crossover child for the next generation.
- **Crossover fraction (CF):** Crossover fraction specifies the fraction of each population, other than elite children, that are made up of crossover children.
- **Selection function:** Selection function specifies how GA chooses parents for the next generation.
- **Migration options:** Migration options determine how individuals move between subpopulations. Migration occurs if the population size is set to be a vector of length greater than 1. When migration occurs, the best individuals from one subpopulation replace the worst individuals in another subpopulation. Individuals that migrate from one subpopulation to another are copied. They are not removed from the source subpopulation.
- **Stopping criteria options:** Stopping criteria options specify the causes of terminating the algorithm.

In this paper, the configuration of GA is set at the values given in Table 3.

| Parameter | Value |
|---|---|
| CF | 0.4,0.6,0.8 |
| PS | 90,200,500 |
| Selection function | Tournament |
| Mutation function | Constraint dependent |
| Crossover function | Intermediate |
| Migration direction | Forward |
| Migration fraction | 0.2 |
| Migration interval | 20 |
| Stopping criteria | Fitness limit to $10^{-4}$ |

Table 3. GA configuration parameters.

Furthermore, the multi-objective optimization of the proposed fuzzy-PID controller is done with respect to six design variables and two objective functions (OFs). The base values $[K_{i\theta}^b, K_{p\theta}^b, K_{d\theta}^b]$ and regulation values $[K_{i\theta}^r, K_{p\theta}^r, K_{d\theta}^r]$ are the design variables. The System's angle of deviation from equilibrium point and the thrust vector's angle of deviation are, respectively, defined as OF1 and OF2:

$$OF1 = \int |\theta| dt \qquad (20)$$

$$OF2 = \int |\Phi| dt \qquad (21)$$

# 5. Results and discussion

In this section, by regarding two aforementioned OFs, the impact of two parameters of PS and CF is measured in the optimization. Figure 4 represents Pareto fronts of these two functions after optimization. Meanwhile, Figure 5 shows the system's position under performance of the designed controller. The angle of thrust vector of the CanSat carrier system is demonstrated in Figure 6. Tables 4-6 display the magnitude of design variables. OF1 and OF2 are shown for optimum points of $A_i$, $B_i$, and $C_i$ in Table 7-9. The best values satisfying the two OFs with the constraints of minimum settling time and overshoot are presented. The relevant magnitude of PS and CF to each figure is brought in its legend. The points $A_i (i = 1,2,...,9)$, $B_i (i = 1,2,...,9)$, and $C_i (i = 1,2,...,9)$ are, respectively, the best for the first, the second, and both OFs.

| Design variable | Point PS=90/CF=0.4 | Value | Point PS=90/CF=0.6 | Value | Point PS=90/CF=0.8 | Value |
|---|---|---|---|---|---|---|
| $K_{i\theta}^b$ | | -0.0094 | | 0.0050 | | 0.013 |
| $K_{\theta}^b$ | | 2.83 | | 2.71 | | 2.93 |
| $K_{d\theta}^b$ | A₁ | 0.36 | A₂ | 0.30 | A₃ | 0.35 |
| $K_{i\theta}^r$ | | 0.36 | | -0.036 | | -0.25 |
| $K_{\theta}^r$ | | 0.46 | | 3.17 | | 0.68 |
| $K_{d\theta}^r$ | | 0.95 | | 1.94 | | 0.83 |
| $K_{i\theta}^b$ | | -0.0075 | | 0.044 | | -0.039 |
| $K_{\theta}^b$ | | -0.0023 | | 0.00019 | | -0.0069 |
| $K_{d\theta}^b$ | B₁ | 2.90 | B₂ | 2.31 | B₃ | 2.11 |
| $K_{i\theta}^r$ | | 0.022 | | 0.021 | | -0.036 |
| $K_{\theta}^r$ | | -0.022 | | -0.26 | | 0.39 |
| $K_{d\theta}^r$ | | 1.68 | | 3.19 | | 1.71 |
| $K_{i\theta}^b$ | | -0.0094 | | 0.033 | | 0.036 |
| $K_{\theta}^b$ | | 2.83 | | 2.48 | | 2.33 |
| $K_{d\theta}^b$ | C₁ | 0.36 | C₂ | 0.31 | C₃ | 0.36 |
| $K_{i\theta}^r$ | | 0.36 | | 0.025 | | 0.044 |
| $K_{\theta}^r$ | | 0.46 | | 3.083 | | 0.34 |
| $K_{d\theta}^r$ | | 0.95 | | 2.05 | | 1.022 |

Table 4. Design variables for $A_i$, $B_i$ and $C_i$ (i =1, 2, 3).

| Design variable | Point PS=200/CF=0.4 | Value | Point PS=200/CF=0.6 | Value | Point PS=200/CF=0.8 | Value |
|---|---|---|---|---|---|---|
| $K_{i\theta}^b$ | | 0.015 | | -0.0025 | | -0.01061 |
| $K_{\theta}^b$ | | 2.79 | | 2.59 | | 2.7385 |
| $K_{d\theta}^b$ | $A_4$ | 0.35 | $A_5$ | 0.33 | $A_6$ | 0.3856 |
| $K_{i\theta}^r$ | | -0.60 | | 0.13 | | 0.4005 |
| $K_{\theta}^r$ | | 1.76 | | 0.63 | | 0.8030 |
| $K_{d\theta}^r$ | | 0.92 | | -0.17 | | -1.1440 |
| $K_{i\theta}^b$ | | -0.088 | | -0.15 | | 0.01091 |
| $K_{\theta}^b$ | | -0.00020 | | 0.0017 | | 0.000345 |
| $K_{d\theta}^b$ | $B_4$ | 2.84 | $B_5$ | 2.48 | $B_6$ | 3.2541 |
| $K_{i\theta}^r$ | | -0.0046 | | 0.055 | | 0.009406 |
| $K_{\theta}^r$ | | 0.45 | | 0.64 | | -0.08872 |
| $K_{d\theta}^r$ | | 2.97 | | 0.78 | | 1.5818 |
| $K_{i\theta}^b$ | | 0.093 | | 0.016 | | 0.009348 |
| $K_{\theta}^b$ | | 2.75 | | 2.23 | | 2.7109 |
| $K_{d\theta}^b$ | $C_4$ | 0.40 | $C_5$ | 0.33 | $C_6$ | 0.3817 |
| $K_{i\theta}^r$ | | 0.056 | | 0.16 | | 0.3722 |
| $K_{\theta}^r$ | | 1.94 | | 0.68 | | 0.5992 |
| $K_{d\theta}^r$ | | 0.53 | | -0.055 | | -0.4311 |

Table 5. Design variables for $A_i$, $B_i$ and $C_i$ (i =4, 5, 6).

| Design variable | Point PS=500/CF=0.4 | Value | Point PS=500/CF=0.6 | Value | Point PS=500/CF=0.8 | Value |
|---|---|---|---|---|---|---|
| $K_{i\theta}^b$ | | 0.0018 | $A_8$ | -0.013 | $A_9$ | -0.014 |
| $K_{\theta}^b$ | | 2.55 | | 2.56 | | 2.71 |
| $K_{d\theta}^b$ | $A_7$ | 0.31 | | 0.35 | | 0.36 |
| $K_{i\theta}^r$ | | 0.040 | | 0.48 | | 0.61 |
| $K_{\theta}^r$ | | 0.52 | | 1.30 | | 1.29 |
| $K_{d\theta}^r$ | | 1.50 | | 0.32 | | -0.57 |
| $K_{i\theta}^b$ | | -0.011 | $B_8$ | -0.17 | $B_9$ | -0.040 |
| $K_{\theta}^b$ | | 0.00013 | | 0.030 | | 0.0064 |
| $K_{d\theta}^b$ | $B_7$ | 2.43 | | 1.26 | | 0.77 |
| $K_{i\theta}^r$ | | -0.0015 | | 0.027 | | 0.012 |
| $K_{\theta}^r$ | | 0.058 | | 0.86 | | 0.16 |
| $K_{d\theta}^r$ | | 2.84 | | 1.42 | | 0.36 |
| $K_{i\theta}^b$ | | 0.0017 | $C_8$ | -0.013 | $C_9$ | -0.00073 |
| $K_{\theta}^b$ | | 2.55 | | 2.56 | | 2.60 |
| $K_{d\theta}^b$ | $C_7$ | 0.31 | | 0.35 | | 0.38 |
| $K_{i\theta}^r$ | | 0.040 | | 0.48 | | 0.52 |
| $K_{\theta}^r$ | | 0.52 | | 1.30 | | 0.54 |
| $K_{d\theta}^r$ | | 1.50 | | 0.32 | | -0.45 |

Table 6. Design variables for $A_i$, $B_i$ and $C_i$ (i =7, 8, 9).

| Objective Function | Point PS=90/CF=0.4 | Value | Point PS=90/CF=0.6 | Value | Point PS=90/CF=0.8 | Value |
|---|---|---|---|---|---|---|
| OF1 | $A_1$ | 0.029 | $A_2$ | 0.030 | $A_3$ | 0.030 |
| OF2 | | 0.042 | | 0.045 | | 0.045 |
| OF1 | $B_1$ | 0.40 | $B_2$ | 0.40 | $B_3$ | 0.40 |
| OF2 | | 0.000017 | | 0.000011 | | 0.000079 |
| OF1 | $C_1$ | 0.029 | $C_2$ | 0.030 | $C_3$ | 0.000079 |
| OF2 | | 0.042 | | 0.039 | | 0.034 |

Table 7. Objective functions for $A_i$, $B_i$ and $C_i$ (i =1, 2, 3).

| Objective Function | Point PS=200/CF=0.4 | Value | Point PS=200/CF=0.6 | Value | Point PS=200/CF=0.8 | Value |
|---|---|---|---|---|---|---|
| OF1 | $A_4$ | 0.029 | $A_5$ | 0.030 | $A_6$ | 0.029 |
| OF2 | | 0.042 | | 0.042 | | 0.040 |
| OF1 | $B_4$ | 0.40 | $B_5$ | 0.40 | $B_6$ | 0.40 |
| OF2 | | 0.0000037 | | 0.000035 | | 0.0000073 |
| OF1 | $C_4$ | 0.032 | $C_5$ | 0.032 | $C_6$ | 0.030 |
| OF2 | | 0.037 | | 0.0359 | | 0.039 |

Table 8. Objective functions for $A_i$, $B_i$ and $C_i$ (i =4, 5, 6).

| Objective Function | Point PS=500/CF=0.4 | Value | Point PS=500/CF=0.6 | Value | Point PS=500/CF=0.8 | Value |
|---|---|---|---|---|---|---|
| OF1 | $A_7$ | 0.030 | $A_8$ | 0.030 | $A_9$ | 0.029 |
| OF2 | | 0.040 | | 0.038 | | 0.041 |
| OF1 | $B_7$ | 0.40 | $B_8$ | 0.39 | $B_9$ | 0.40 |
| OF2 | | 0.0000021 | | 0.00012 | | 0.000065 |
| OF1 | $C_7$ | 0.030 | $C_8$ | 0.030 | $C_9$ | 0.030 |
| OF2 | | 0.040 | | 0.038 | | 0.037 |

Table 9. Objective functions for $A_i$, $B_i$ and $C_i$ (i =7, 8, 9).

Further, as seen in Figure 4, point $A_i$ and $C_i$ are in a near proximity in which in some cases a coincidence occurs. It is mainly due to non-convergence of CanSat carrier launch vehicle points with points far from $A_i$. A similar behaviour is observed from Pareto fronts of the situation, angular velocity, and angle of the thrust vector for the launch vehicle.

To analyse the impact of each parameter in GA, Figures 7-12 are produced. Figure 7 shows the dependency of OF1 (at points $A_i$) to nine different combination forms of GA parameters. The figure shows that the minimum area under "situation of launch vehicle" curve is obtainable for PS=200 and CF=0.8. It is also inferred that for better results, parameters PS and CF must be increased simultaneously. For low PS, increasing CF helps to improve first OF, but with more magnitudes of PS, higher CFs yield better results.

Figure 8 represents dependency of OF2 (at points $A_i$) to nine different forms of combinations of GA parameters. The figure shows the least area below the deviation angle of the thrust curve for CanSat carrier system when PS=500 and CF=0.6. Figures 9 and 10 propose that the smallest magnitude for the first and second OFs (pertaining to $B_i$) is achievable for, respectively, PS=500 and CF=0.6 and PS=500 and CF=0.4. In Figures 11 and 12, magnitudes of the first and second objective functions in $C_i$ Points are represented, respectively. The first OF proposed the point $C_1$ with PS=90 and CF=0.4. Meanwhile, the second function insists on the point with PS=90 and CF=0.8.

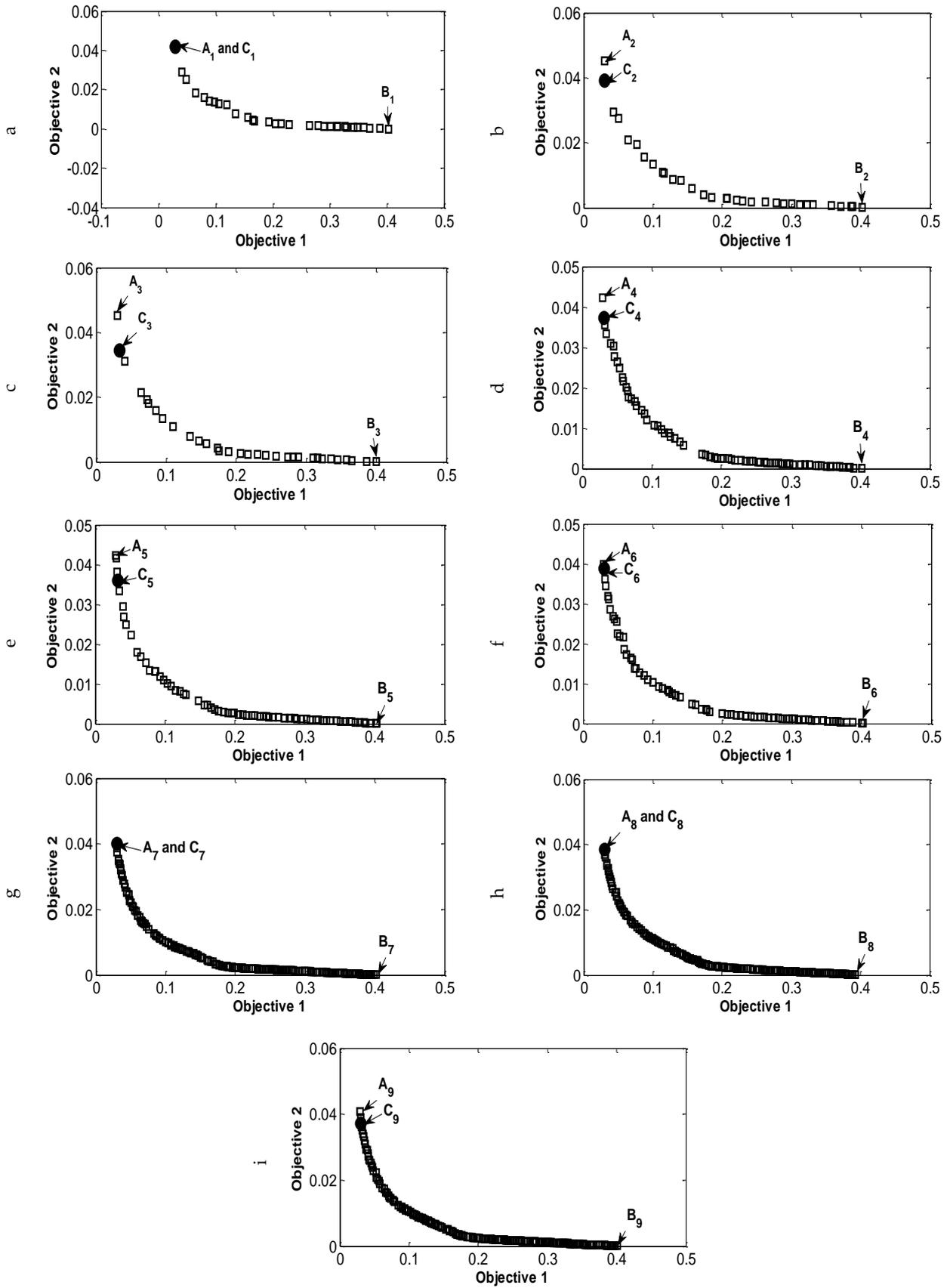

Figure 4. Pareto front by Objectives 1 and 2 corresponds to the (a) PS=90 & CF=0.4 (b) PS=90 & CF=0.6 (c) PS=90 & CF=0.8 (d) PS=200 & CF=0.4 (e) PS=200 & CF=0.6 (f) PS=200 & CF=0.8 (g) PS=200 & CF=0.4 (h) PS=200 & CF=0.6 (i) PS=200 & CF=0.8

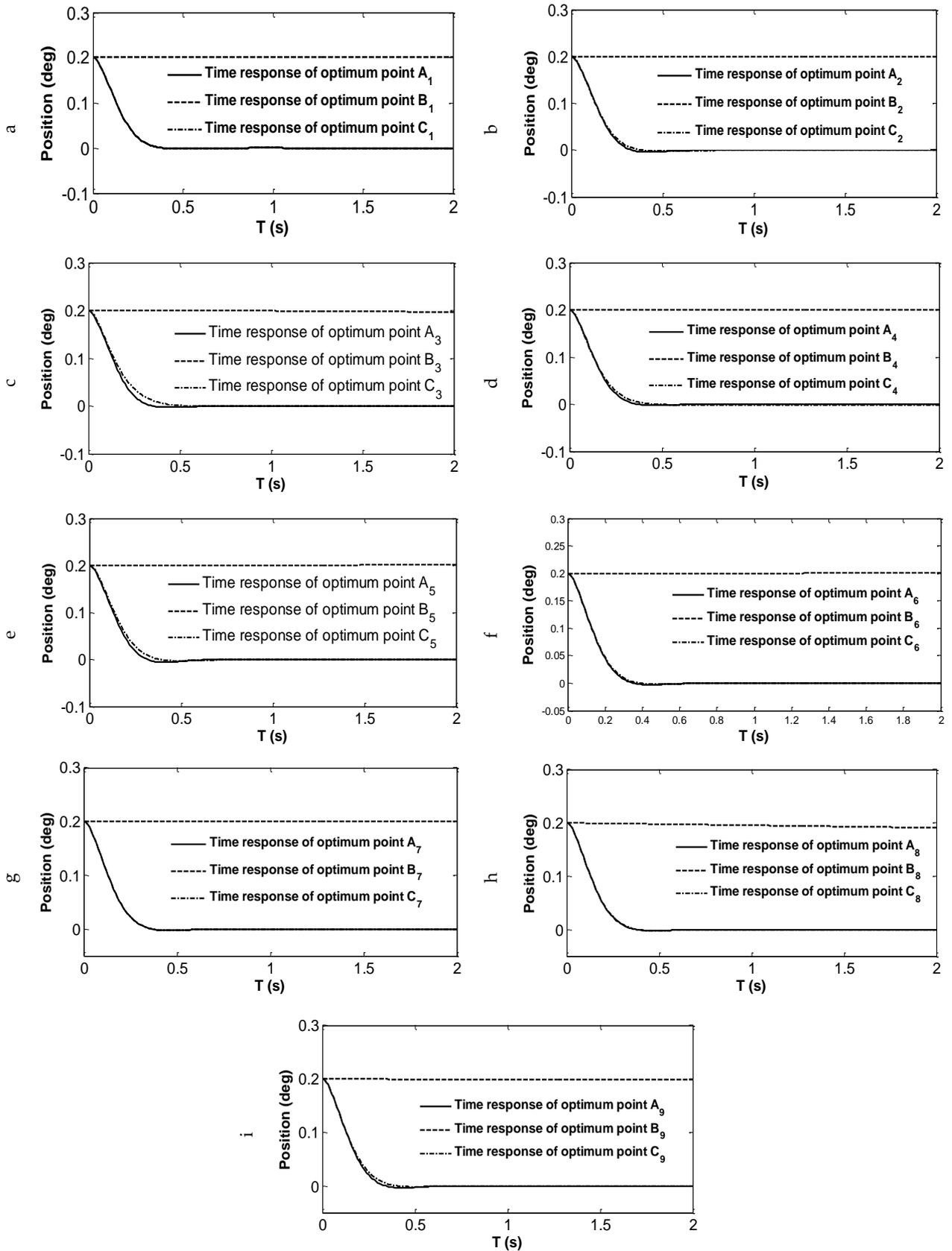

Figure 5. Time response of the CanSat carrier system's position for (a) $A_1$, $B_1$, and $C_1$ (b) $A_2$, $B_2$, and $C_2$, (c) $A_3$, $B_3$, and $C_3$ (d) $A_4$, $B_4$, and $C_4$ (e) $A_5$, $B_5$, and $C_5$ (f) $A_6$, $B_6$, and $C_6$ (g) $A_7$, $B_7$, and $C_7$ (h) $A_8$, $B_8$, and $C_8$ (i) $A_9$, $B_9$, and $C_9$ as the optimum points.

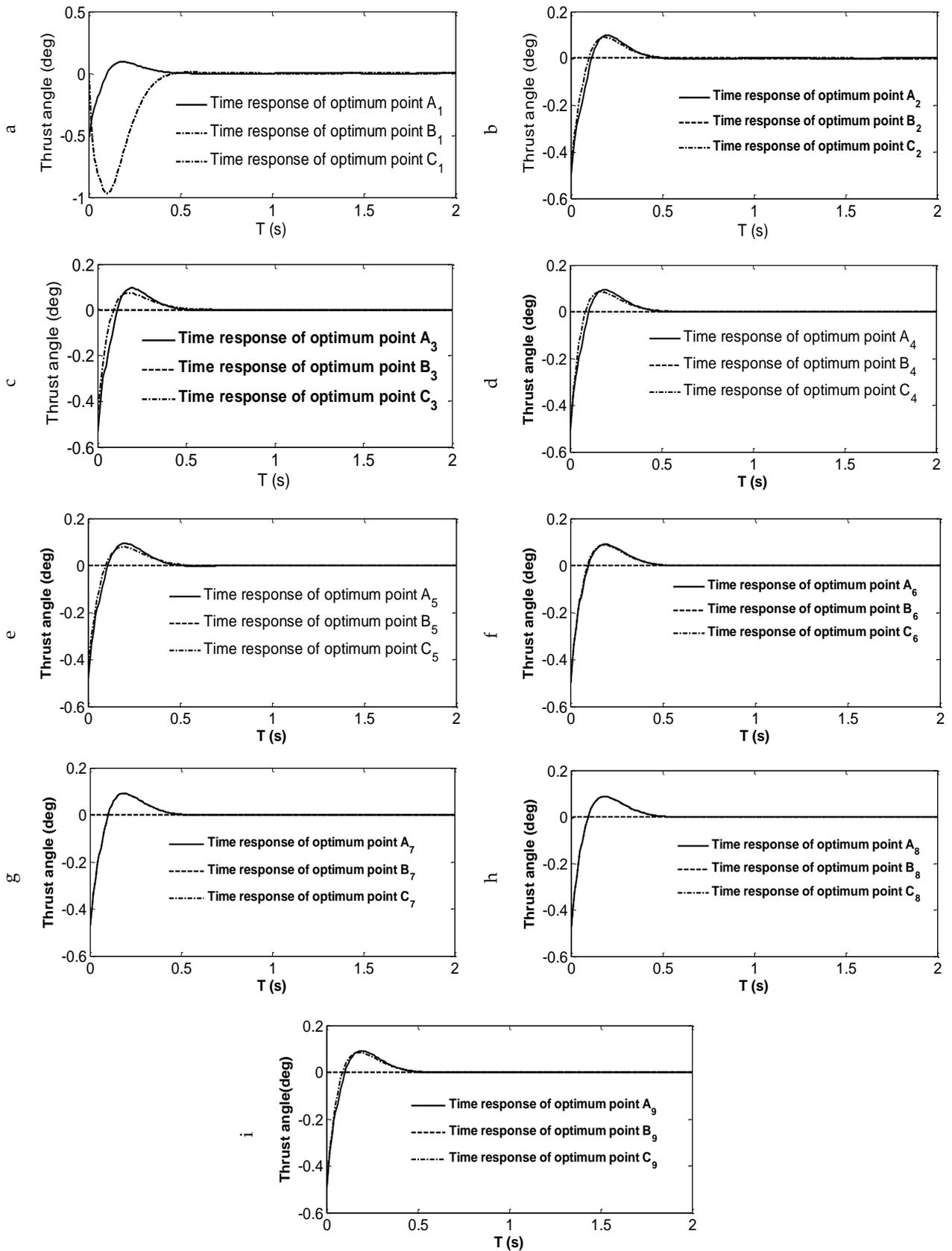

Figure 6. Time response of the thrust angle of CanSat carrier system for (a) $A_1$, $B_1$, and $C_1$ (b) $A_2$, $B_2$, and $C_2$, (c) $A_3$, $B_3$, and $C_3$ (d) $A_4$, $B_4$, and $C_4$ (e) $A_5$, $B_5$, and $C_5$ (f) $A_6$, $B_6$, and $C_6$ (g) $A_7$, $B_7$, and $C_7$ (h) $A_8$, $B_8$, and $C_8$ (i) $A_9$, $B_9$, and $C_9$ as the optimum points.

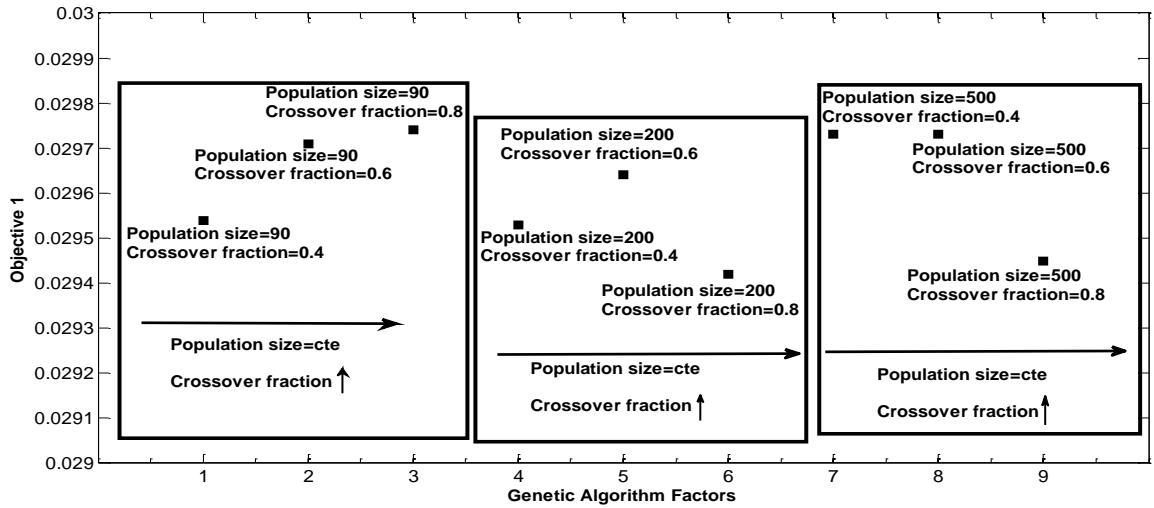

Figure 7. GA parameters versus OF1 for the best points from the viewpoint of OF1.

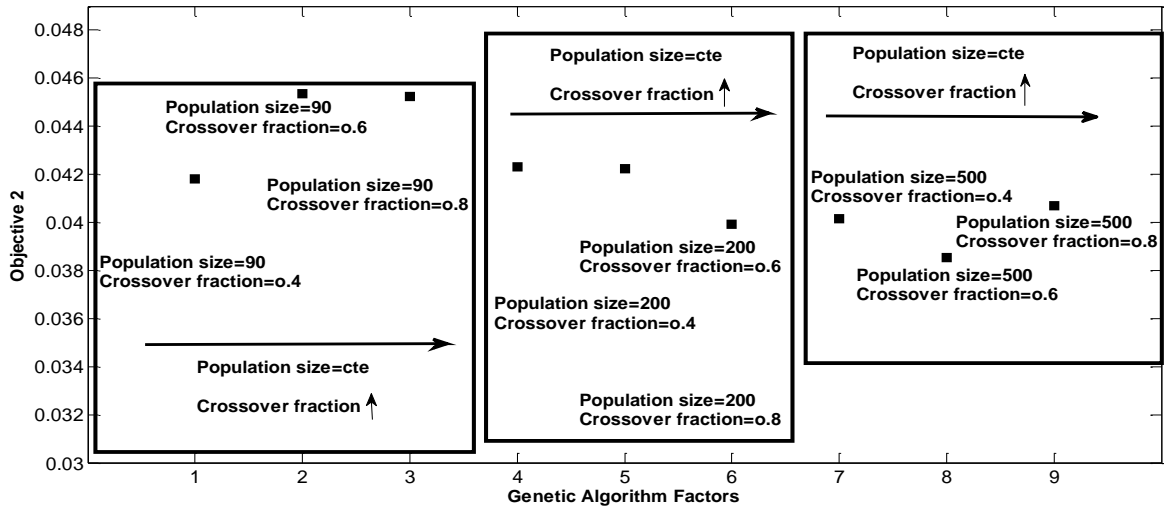

Figure 8. GA parameters versus OF2 for the best points from the viewpoint of OF1.

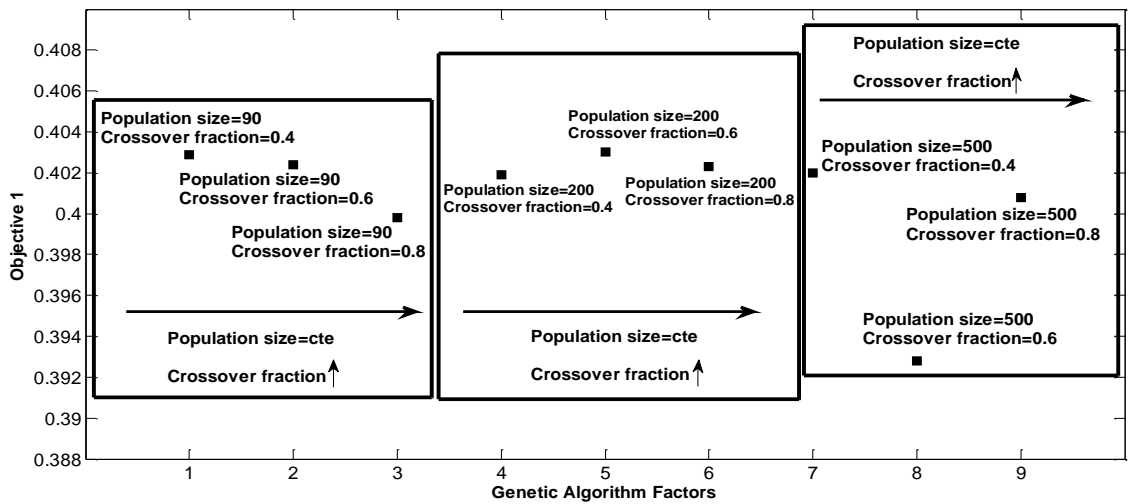

Figure 9. GA parameters versus OF1 for the best points from the viewpoint of OF2.

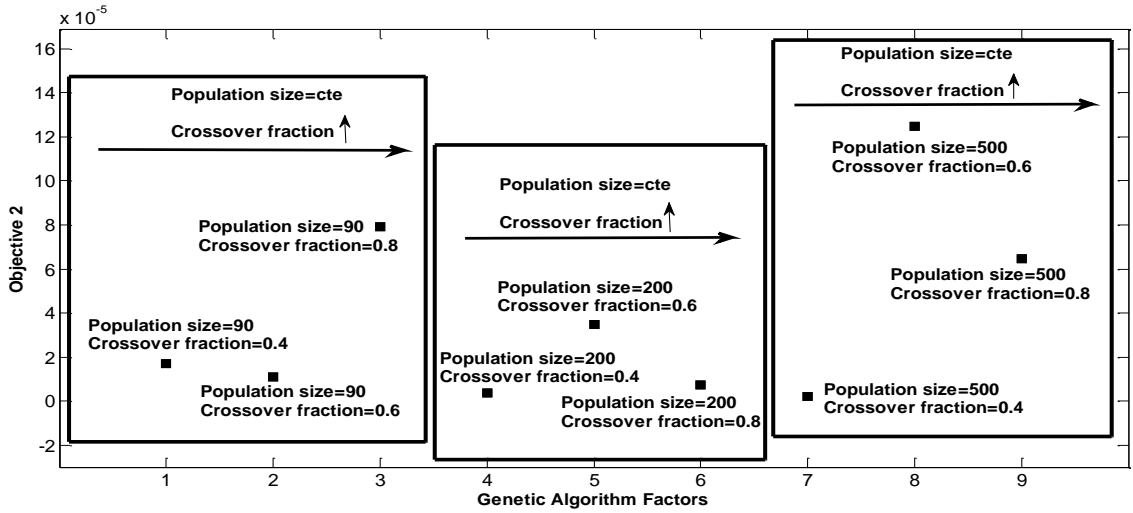

Figure 10. GA parameters versus OF2 for the best points from the viewpoint of OF2.

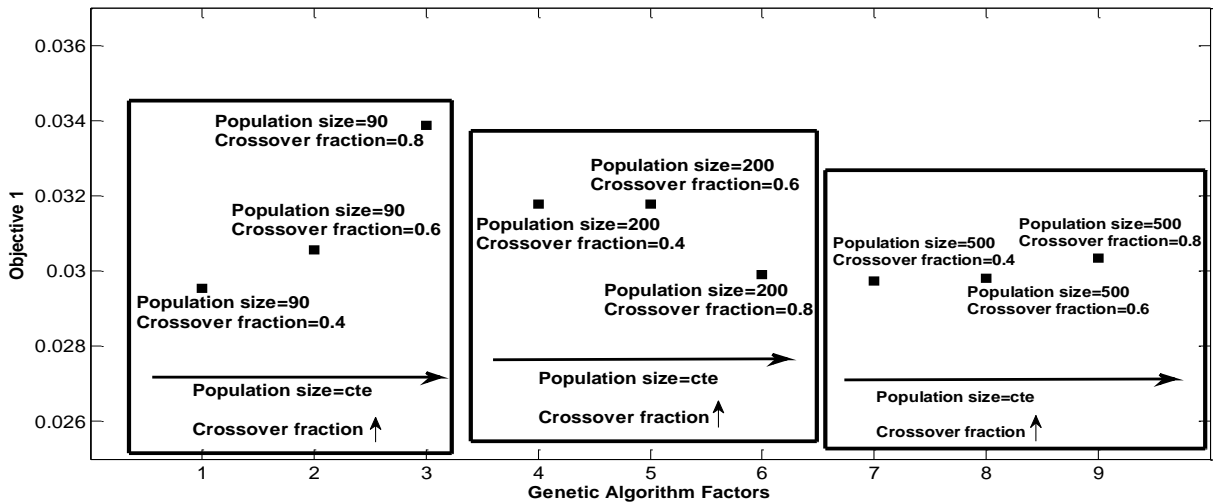

Figure 11. GA parameters versus OF1 for the best points from the viewpoint of OF1 and OF2.

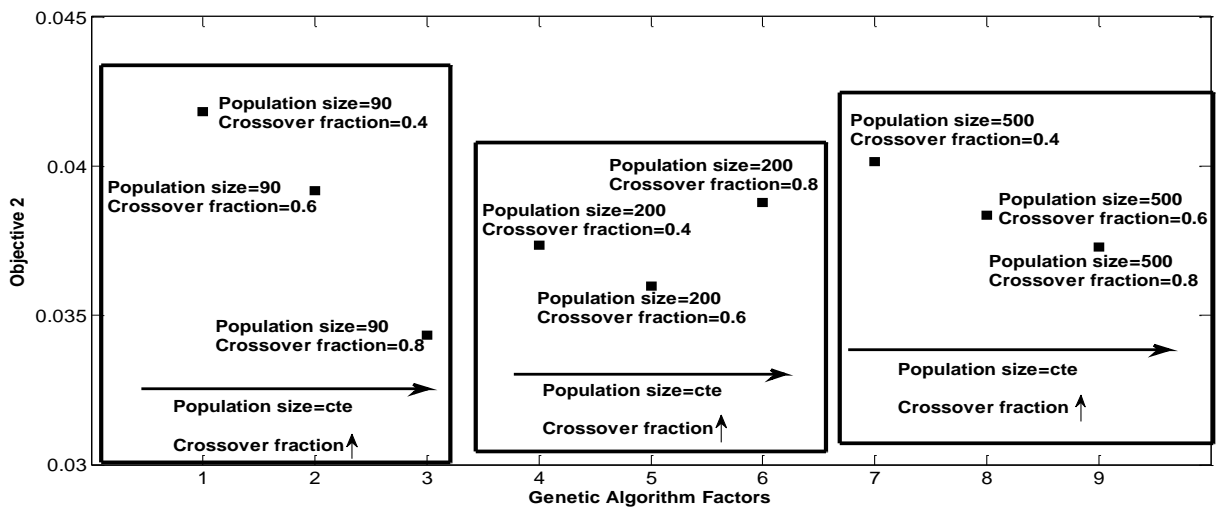

Figure 12. GA parameters versus OF2 for the best points from the viewpoint of OF1 and OF2.

## 6. Conclusion

This paper represented a design of a fuzzy controller based on the GA for the purpose of controlling thrust vector of a launch vehicle which carries CanSat. Minimizing the errors caused by system deviation from equilibrium state and thrust vector deviation are two objectives for optimizing this controller. This was done by manipulating GA parameters in nine different combination forms to satisfy each objective function and also both of them simultaneously. Further it was examined how these parameters affect the optimal points. By observing constraints of minimum settling time and overshoot, the results showed that the optimal points proposed by the first OF are in proximity with the ones from both OFs which in some cases end in coincidence. Finally, by comparing magnitudes of OFs for various combinations of GA parameters, the optimum points and their relevant parameters were introduced.

## Conflict of Interest

The authors declare that there is no conflict of interest.